# Closed-Loop Sim-to-Real Reinforcement Learning for Deformable Microfiber Shape Control

Alessandro Amici, Houari Bettahar, Veeti Jaakkola, Quan Zhou

*Abstract*— **Autonomous contact-based micromanipulation is challenging because surface and interfacial interactions at the microscale are difficult to model accurately, limiting the use of conventional model-based control and sim-to-real learning. We present a closed-loop sim-to-real reinforcement learning (RL) approach for microfiber shape control on a surface. The central idea is to train geometric shape regulation in a simplified frictionless simulator and rely on real-time visual feedback during deployment to iteratively correct the observed effects of unmodeled surface interactions. An RL policy trained entirely in simulation is transferred directly to a physical dual-gripper micromanipulation system operating at 40 Hz, without retraining or domain adaptation. Using silk microfibers as a testbed, the policy achieves a mean point-wise shape error of 270 ± 80 µm across twenty-four diverse initial configurations. Across nine specimens covering all combinations of three fiber diameters (50, 80, and 120 µm) and three manipulated lengths (10 mm, 15mm, and 20 mm), the same policy achieves sub-millimeter final shape error without any retraining or retuning. These results show that a policy learned in a simplified simulator can achieve repeatable real-world microfiber shape regulation under surface contact, provided that the task-relevant effects of the sim-to-real mismatch remain observable and correctable within the closed feedback loop.**

## I. Introduction

Robotic micromanipulation is a key technology enabling the precise handling of micrometer-scale objects for applications spanning biological cell studies [1], [2], minimally invasive surgery [3], [4], micro-assembly [5], [6], and fiber-optic alignment [7], [8]. Despite substantial advances in micromanipulation systems, modeling, and control strategies, the achievable level of autonomy remains limited, and in practice most contact-based micromanipulation systems still rely on teleoperation or heuristic, task-specific controllers [9]. A core difficulty is that the governing physics change at the microscale: gravity and inertia become negligible relative to surface and interfacial interactions, particularly adhesion-related effects arising from capillary, van der Waals, and electrostatic forces, as well as distributed frictional contact. These interactions are often history-dependent and difficult to model accurately [10], [11], which makes it challenging to design reliable and generalizable control policies. Consequently, achieving autonomy in robotic micromanipulation remains a significant challenge, despite growing interests in learning-based approaches [12].

By comparison, at the macroscale, reinforcement learning (RL) has demonstrated compelling results in autonomous robotic control, from contact-rich in-hand manipulation of complex objects such as a Rubik's cube [13] to champion-level drone racing [14]. One central enabler of these successes has been sim-to-real transfer [15]: by training entirely in simulation, RL agents can acquire complex manipulation skills at scale, without the prohibitive cost, time, and complexity of collecting real-world experience [16].

This work was supported by the Finnish Ministry of Education and Culture through the Intelligent Work Machines Doctoral Education Pilot Program (IWM VN/3137/2024-OKM-4), and the Research Council of Finland grant 362715. *(Corresponding authors: Houari Bettahar, Quan Zhou)*

Amici Alessandro, Houari Bettahar, Jaakkola Veeti and Quan Zhou are with the Department of Electrical Engineering and Automation, Aalto University, 02150 Espoo, Finland (e-mail: firstname.lastname@aalto.fi).

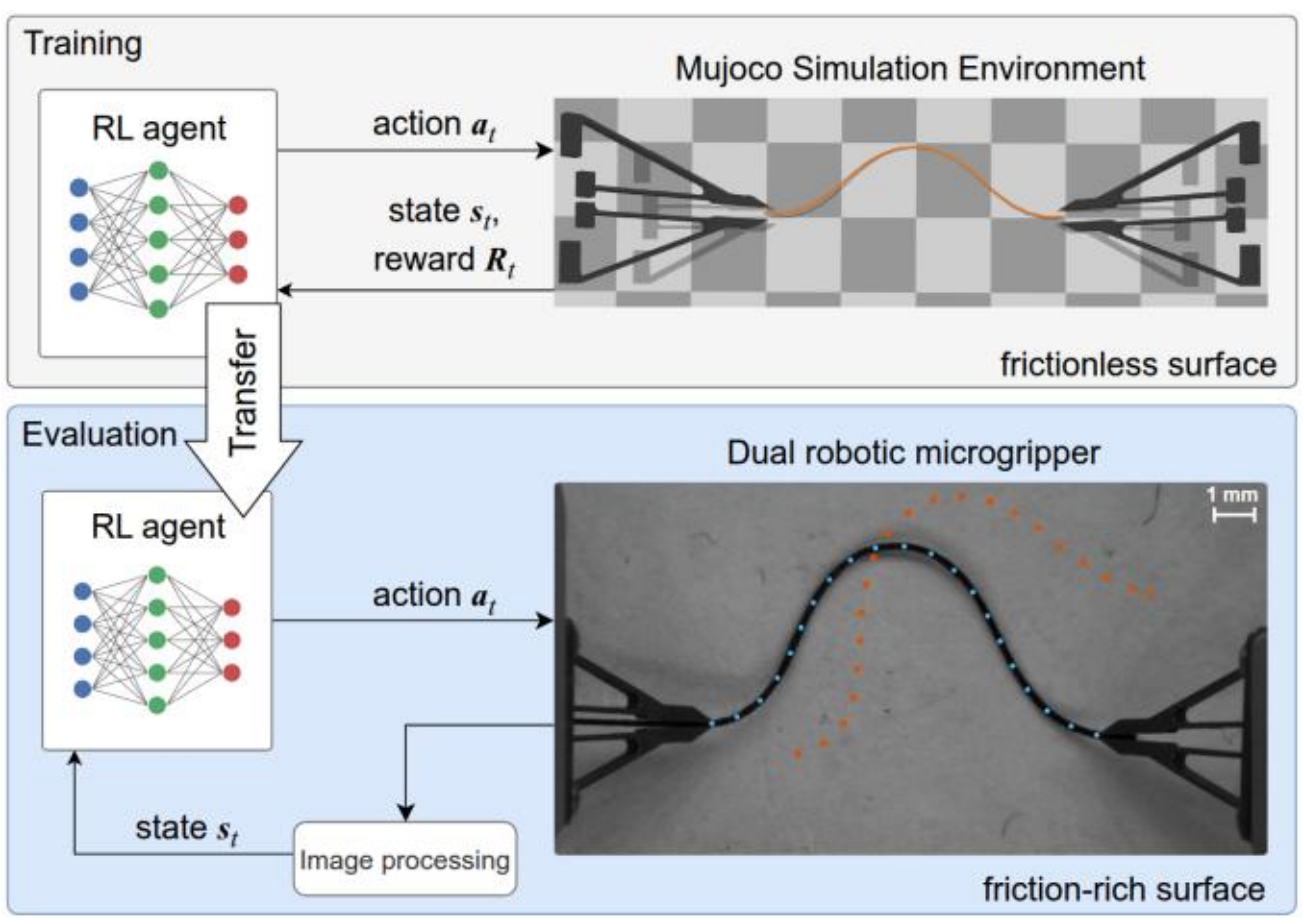

**Fig. 1:** Overview of the proposed deformable microfiber shape control approach. An RL agent is trained in a simplified frictionless MuJoCo simulator and deployed directly on a physical dual-gripper micromanipulation system, where it shapes real microfibers under surface contact via closed-loop visual-feedback control.

Extending this paradigm to the microscale is attractive, but substantially more challenging because the dominant interactions are harder to capture accurately in simulation.

Recent works have begun to explore sim-to-real RL for contact micromanipulation. Gong *et al.* [17] studied automatic cell rotation using deep Q-learning but evaluated the method only in simulation. Likewise, Zhang *et al.* [18] applied PPO and A2C to posture adjustment of zebrafish embryonic cells in PyBullet [19], but did not report transfer to physical experiments. Related sim-to-real RL studies was also applied at slightly larger scales, for example in millimeter-scale rigid insertion [20] and deformable rope cutting [21]. These studies indicate that sim-to-real RL for contact-rich robotic micromanipulation remains underexplored, particularly for deformable objects.

In this paper, we propose a closed-loop sim-to-real reinforcement learning approach for deformable microfiber shape control on a surface, as illustrated in Fig. 1. The central idea is to separate geometric skill learning from accurate microscale contact modeling: the policy is trained in a simplified frictionless simulator to learn endpoint-based shape regulation, and is then deployed directly on the physical system, where real-time visual feedback at 40 Hz allows the controller to iteratively correct the effect of unmodeled surface interactions. We implement this approach on a dual robotic microgripper-based micromanipulation system that shapes microfibers by independently controlling both endpoints. Despite the simplified training environment, the learned policy achieves sub-millimeter shape control accuracy across nine fiber specimens of varying diameters and lengths and generalizes to diverse initial configurations without retraining or domain adaptation.

## II. Problem Formulation

In this work, we use microfiber shaping as a testbed for studying sim-to-real transfer in microscale deformable manipulation. This task captures key challenges of contact-based manipulation arising from endpoint actuation, object compliance, and persistent surface contact. As a result, the fiber shape is governed not only by the gripper-imposed boundary conditions, but also by the fiber's internal mechanics and history-dependent distributed surface interactions. At the microscale, gravity and inertia are negligible relative to adhesive and frictional contact effects, making the mapping from gripper motion to fiber shape difficult to model accurately. This makes microfiber shaping significantly different from a purely geometric shaping task and motivates its formulation as a feedback-driven control problem. We therefore approach the task within a reinforcement learning framework, using closed-loop feedback to regulate the observed fiber shape.

We formalize this task by considering the shape control of a deformable microfiber resting on a planar surface. The fiber has a total length of $l$ and a diameter of $d$ and is composed of a flexible elastic material. The fiber is held at both endpoints by two robotic grippers, denoted L (left) and R (right). The positions of left and right grippers at time $t$ are denoted $\boldsymbol{x}_t = (\boldsymbol{x}_{\mathrm{L},t}, \boldsymbol{x}_{\mathrm{R},t}) \in \mathbb{R}^4$, and constitute the sole controllable inputs to the system. Both grippers move independently in the $xy$-plane. The $z$-axis is controllable; however, it is kept fixed during experiments to maintain stable contact between the fiber and the surface. Under these constraints, the planar fiber configuration evolves as a function of endpoint motion, fiber mechanics, and surface interactions.

We describe the configuration of the fiber at time $t$ by its planar centerline $\mathcal{C}_t$, represented as an ordered sequence of $N$ discrete points uniformly sampled along the fiber:

$$\mathcal{C}_t = \{\boldsymbol{q}_{i,t} \in \mathbb{R}^2 \mid i = 1, \ldots, N\} \tag{1}$$

with boundary conditions $\boldsymbol{q}_{1,t} = \boldsymbol{x}_{\mathrm{L},t}$ and $\boldsymbol{q}_{N,t} = \boldsymbol{x}_{\mathrm{R},t}$ imposed by the grippers. The target shape $\mathcal{C}^*$ is selected and the initial endpoint configuration $\boldsymbol{x}_0$ is sampled from a feasible set of gripper positions $\mathcal{G}$. The trial begins once the grippers have reached $\boldsymbol{x}_0$, at which point the fiber assumes an initial centerline $\mathcal{C}_0$ determined by the endpoint boundary conditions and the fiber–surface interactions arising during initialization.

The objective is to learn a control policy $\pi: \mathcal{S} \to \mathcal{A}$ that maps the observed system state $\boldsymbol{s}_t \in \mathcal{S}$ to gripper velocity commands $\boldsymbol{a}_t \in \mathcal{A}$ such that the fiber centerline $\mathcal{C}_t$ converges to the target shape $\mathcal{C}^*$. The task is considered successful if the point-wise shape error falls below a predefined tolerance $\epsilon_{\mathrm{tol}}$ within a fixed horizon of $T$ timesteps; otherwise, the trial terminates as a failure.

## III. Methodology

In the proposed method, we train a control policy in a simplified frictionless simulator and transfer it directly to the physical micromanipulation system in closed loop. The deployment strategy relies on repeated visual-feedback correction of the observed fiber shape, as detailed in Section III.C.

### A. Simulation Environment

The simulation environment is deliberately simplified to train the geometric shape-regulation component of the task under stable and controllable conditions. The fiber is modeled in MuJoCo [22] without fiber-surface friction during training, so that the policy learns to regulate fiber shape from geometric state information only, without relying on an explicit contact model. The contact mismatch introduced by this simplification is addressed during closed-loop deployment, as discussed in Section III.C.

The microfiber is represented as an articulated chain of $N = 20$ rigid cylindrical segments connected by ball joints over a total arc length of 15 mm. Bending compliance is introduced by tuning joint stiffness and damping to visually resemble real microfiber mechanics. To improve numerical stability at microscale dimensions, we apply a dimensionless geometric scaling factor of 100, use small stable segment masses, and disable gravity. Simulated positions are converted back to physical units before being used in policy input and reward.

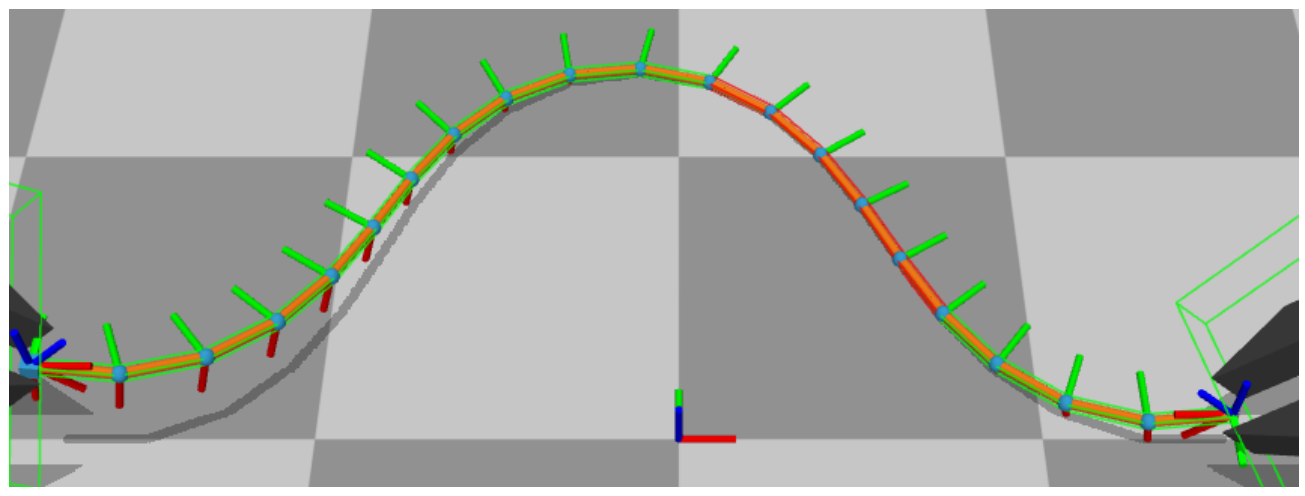

**Fig. 2:** The simulated fiber components. Sphere joints connect the 20 cylinders. The endpoints are welded to the grippers pose. The cylinders encapsulated by a red rectangle are in contact with the frictionless surface, keeping the fiber on the planar surface, but eliminating deformation due to friction.

Fig. 2 shows the rendered simulation environment. Each gripper is modeled as a rigid body with independent prismatic joints along the $x-$, $y-$, and $z-$axes. During training, control is restricted to the planar $xy$-workspace, while the $z$-axis is held fixed. The fiber endpoints are attached to the grippers through equality weld constraints, enforcing the endpoint boundary conditions $\boldsymbol{q}_1 = \boldsymbol{x}_{\mathrm{L}}$ and $\boldsymbol{q}_N = \boldsymbol{x}_{\mathrm{R}}$.

Surface friction is omitted during training because effective fiber–surface friction at the microscale is coupled with adhesion and other interfacial effects, and depends strongly on surface roughness, cleanliness, humidity, and local contact conditions. A simple deterministic friction model would therefore risk introducing simulator-specific contact behavior that does not transfer to the physical system, while a more realistic probabilistic or data-driven contact model would require additional system-identification data beyond the scope of this work. We therefore train with the fiber–surface friction coefficient set to zero and evaluate whether closed-loop visual feedback can mitigate the resulting model mismatch during physical deployment.

### B. Reinforcement learning formulation

The physical microfiber shaping process is only partially observed, since internal stress, interfacial friction properties, and contact history are not measured directly. We therefore treat the memoryless Markov Decision Process (MDP) [23] formulation as a practical approximation for reactive closed-loop control rather than as a complete description of the underlying physical state. Under this assumption, the policy does not require access to the full unobserved interaction history; instead, it can act reactively based on the currently observed fiber geometry and its deviation from the desired target shape. We formulate the task as an MDP $(\mathcal{S}, \mathcal{A}, \mathcal{P}, R, \gamma)$, and instantiate its components as follows:

**State space** ($\mathcal{S} \subset \mathbb{R}^{52}$)**:** At each timestep $t$, the RL agent receives the observed state input $\boldsymbol{s}_t \in \mathcal{S}$, which consists of five components:

$$\boldsymbol{s}_t = [\boldsymbol{x}_t,\ \dot{\boldsymbol{x}}_t,\ \mathcal{C}_t,\ \mathcal{C}^*,\ \boldsymbol{a}_{t-1}] \tag{2}$$

Here, $\boldsymbol{x}_t$ denotes the current planar positions of the left and right grippers, $\dot{\boldsymbol{x}}_t$ their current planar velocities, $\mathcal{C}_t$ the current fiber centerline, $\mathcal{C}^*$ the target centerline and $\boldsymbol{a}_{t-1}$ the previous action. Since the $z$-axis is held fixed, all gripper quantities are restricted to the $xy$-plane. The state vector concatenates the gripper positions and velocities $(\boldsymbol{x}_t,\ \dot{\boldsymbol{x}}_t) \in \mathbb{R}^8$, current and target fiber shapes represented by 10 planar points each $(\mathcal{C}_t,\ \mathcal{C}^*) \in \mathbb{R}^{40}$, and the previous action $(\boldsymbol{a}_{t-1}) \in \mathbb{R}^4$. This yields 52 dimensions for the state vector, providing the policy with the current geometric mismatch, actuator state, and a limited short-horizon actuation history.

**Action space** ($\mathcal{A} = [-1,1]^4$)**:** The agent outputs continuous planar velocity commands for both grippers:

$$\boldsymbol{a}_t = [\dot{x}_L,\ \dot{y}_L,\ \dot{x}_R,\ \dot{y}_R] \in \mathbb{R}^4 \tag{3}$$

Here $(\dot{x}_1, \dot{y}_1)$ and $(\dot{x}_2, \dot{y}_2)$ are the planar velocity commands for the left and right grippers, respectively. The $z$-axis is excluded as it is held fixed during manipulation. Velocity control is used because it provides a natural interface for closed-loop shape regulation and allows the policy to make incremental corrections as the observed fiber shape evolves. The actions are normalized to $[-1,1]^4$ and scaled to physical velocity limits at execution time.

**Reward function** ($R: \mathcal{S} \times \mathcal{A} \to \mathbb{R}$): The reward combines shape tracking, tension avoidance, action smoothness, and a success bonus:

$$R_t = -\, w_{\mathrm{er}}\epsilon_t - w_{\mathrm{te}}P_{\mathrm{te},t} - w_{\mathrm{sm}}P_{\mathrm{sm},t} + B_t \tag{4}$$

where $\epsilon_t$, $P_{\mathrm{te},t}$, $P_{\mathrm{sm},t}$, and $B_t$ denote the shape error, tension penalty, smoothness penalty, and success bonus, respectively, and $w_{\mathrm{er}}, w_{\mathrm{te}}, w_{\mathrm{sm}}$ are the corresponding weights. The shape error combines the mean and worst-case pointwise deviations:

$$\epsilon_t = \frac{e^2_{\mathrm{mean},t} + e^2_{\mathrm{max},t}}{2} \tag{5}$$

$$e^2_{\mathrm{mean},t} = \frac{1}{N}\sum_{i=1}^{N} \|\boldsymbol{q}_{i,t} - \boldsymbol{q}_i^*\|_2^2, \tag{6}$$

$$e^2_{\mathrm{max},t} = \max_i \|\boldsymbol{q}_{i,t} - \boldsymbol{q}_i^*\|_2^2 \tag{7}$$

The tension penalty $P_{\mathrm{te},t}$ is activated when the gripper separation $d_{\mathrm{grip},t} = \|\boldsymbol{x}_{\mathrm{L},t} - \boldsymbol{x}_{\mathrm{R},t}\|_2$ exceeds the safe elongation limit $l_{\max} = \rho l$, with $\rho$ denoting the fraction of fiber length $l$:

$$P_{\mathrm{te},t} = \begin{cases} (d_{\mathrm{grip},t} - l_{\max}), & \text{if } d_{\mathrm{grip},t} > l_{\max} \\ 0, & \text{otherwise} \end{cases} \tag{8}$$

The smoothness penalty is:

$$P_{\mathrm{sm},t} = \|\boldsymbol{a}_t - \boldsymbol{a}_{t-1}\|_2 \tag{9}$$

Finally, the success bonus $B_t > 0$ is awarded when the shape error falls below the success threshold, $\epsilon_t < \epsilon_{\mathrm{thr}}$, and is maintained for 2 s, encouraging the policy to hold the target shape stably.

**Policy and training:** The policy $\pi_\theta: \mathbb{R}^{52} \to \mathbb{R}^4$ is implemented as a feedforward neural network with three hidden layers of 128, 128, and 64 neurons, using ReLU activations. It is trained with PPO [24] in Stable-Baselines3 v2.7 [25] with 30 parallel environments. We use a feedforward policy to test whether reactive shape regulation is sufficient without recurrent state estimation, consistent with the short-impact-horizon assumption discussed in III.C. Standard PPO hyperparameters are listed in Table I .

TABLE I
POLICY ARCHITECTURE AND PPO HYPERPARAMETERS

| Parameter | Value | Parameter | Value |
|---|---|---|---|
| Policy type | Feedforward | Parallel envs. | 30 |
| Input dimension | 52 | Total steps | 20 M |
| Output dimension | 4 | LR schedule | Cosine decay |
| Hidden layers | [128, 128, 64] | Initial LR | $6 \times 10^{-4}$ |
| Activation | ReLU | Final LR | $2 \times 10^{-4}$ |
| Discount, $\gamma$ | 0.99 | GAE, $\lambda$GAE | 0.95 |
| Clip range, $\varepsilon$ | 0.2 | Entropy coef. | 0.01 |
| Batch size | 1024 | Epochs/rollout | 10 |

Training episodes are sampled from a dataset of 6037 geometrically valid fiber configurations generated by sweeping the two gripper positions in simulation and retaining configurations with inter-gripper distance $0.5\, l \le d \le 0.9\, l$, thereby excluding numerically unstable collapsed configurations and near-stretched configurations with limited shape variation. At the start of each episode, the initial and target shapes are sampled independently from this dataset, so the policy learns regulation across diverse initial-target pairs.

Training uses curriculum learning [26], [27] on the success threshold $\epsilon_{thr}$, which is progressively tightened from 1.2 mm to 0.01 mm over 13 levels. The curriculum level is adjusted using the recent success rate over a 50-episode window, increasing when the success rate exceeds 70% and decreasing when it falls below 30%. Each episode is capped at 8 s, and the success bonus $B_t$ is scaled with curriculum level to encourage precise shape regulation.

### C. *Sim-to-real transfer via closed-loop control*

After training, the policy $\pi_\theta$ is deployed directly on the physical dual robotic microgripper system without retraining or fine-tuning. At each step of the 40 Hz control loop, the image-derived fiber centerline $\mathcal{C}_t$ is combined with the gripper state, the target shape $\mathcal{C}^*$, and the previous action $\boldsymbol{a}_{t-1}$ to form the policy input $\boldsymbol{s}_t$. This input is normalized using the running statistics collected during training, and $\pi_\theta$ outputs the planar velocity command $\boldsymbol{a}_t$ for the two grippers.

The methodological assumption is that, in the tested regime, the unmodeled contact dynamics have a sufficiently short impact horizon that their effects can be corrected through repeated feedback updates. In other words, the policy is not required to predict the full hidden contact state or reconstruct long contact histories; it reacts to the geometric deviations that these hidden effects induce in the observed fiber shape. This is consistent with a closed-loop output-regulation view of the task and with the broader sim-to-real RL view that feedback policies can tolerate model mismatch when the task-relevant effects of the mismatch are observed within the control loop [28]. Under this view, iterative feedback correction can remain effective even when the underlying fiber-surface contact dynamics are not modeled explicitly, provided that the dominant mismatch appears quickly in the measured output and remains recoverable through subsequent actions.

### D. Fiber centerline estimation

In simulation, the fiber centerline $\mathcal{C}_t$ is obtained directly from the MuJoCo state by reading the planar positions of the fiber segments. In the physical system, by contrast, the centerline is not directly measurable and must be estimated from camera images. This introduces an additional observation mismatch, since the policy is trained on noise-free simulator state but deployed using image-derived states affected by detection noise and calibration error.

At each control step, a top-view image is cropped to a region of interest excluding the gripper bodies, converted to grayscale, smoothed, and processed using OpenCV's Canny edge detection [29], [30]. The extracted fiber contour is then resampled to a fixed number of points along arc length and transformed to the global coordinate frame using a pre-calibrated homography transformation. The resulting point set provides the centerline representation $\mathcal{C}_t$, which is used to construct the state vector $\boldsymbol{s}_t$and compute the shape error $\epsilon_t$. Algorithm 1 summarizes the complete method.

## IV. Experimental results and discussions

### A. Experimental setup

The experimental platform consists of two opposing microgripper assemblies for fiber manipulation (Fig. 3). Each assembly comprises a microgripper (SmarAct, SG06-37) mounted at a 45° angle on a 3-DOF precision motorized stage (SmarAct, XYZ:17-22). During autonomous operation, only planar motion is actively controlled, while the z-axis is held fixed to maintain surface contact. The effective manipulation workspace is approximately 23 × 12.8 × 12.8 mm. Fiber state estimation and autonomous control are performed using a top-view camera (FLIR, Grasshopper3 GS3-U3-51S5C) equipped with a long-working-distance objective (Infinity, InfiniMite Alpha). A side-view camera of the same model is used for operator monitoring. Uniform illumination is provided by a ring light and auxiliary side lighting to reduce shadows and improve contour detection.

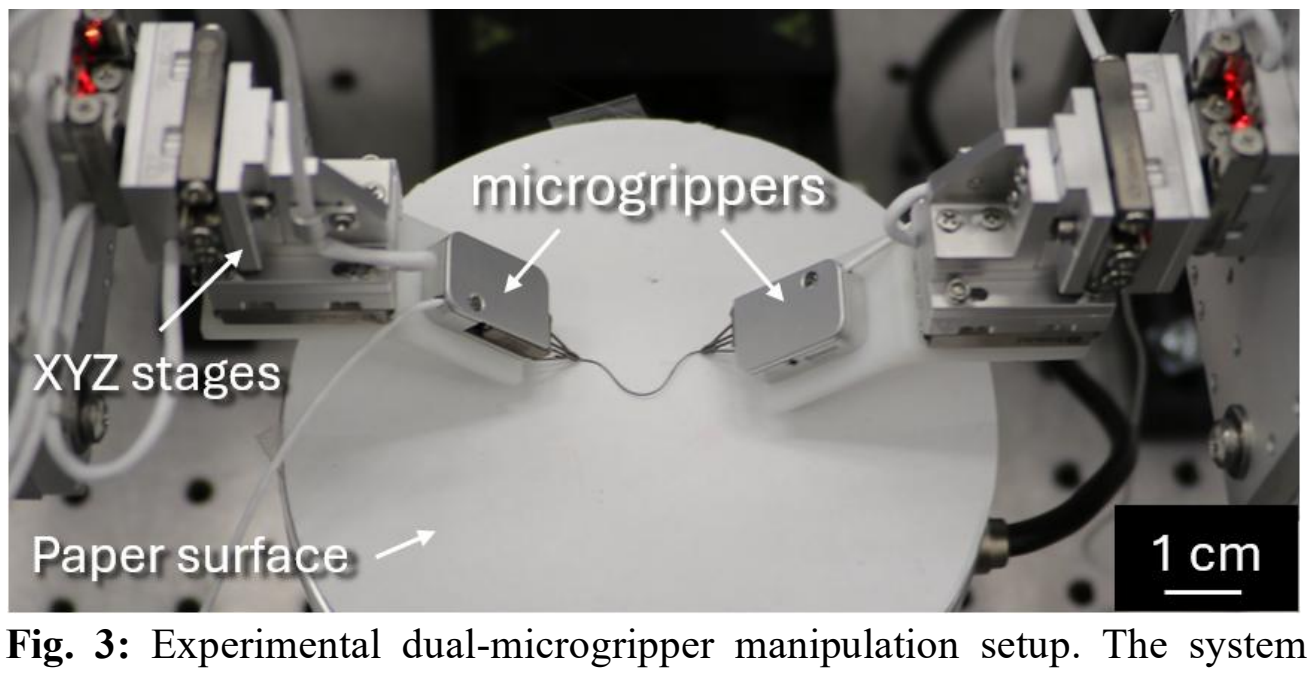

**Fig. 3:** Experimental dual-microgripper manipulation setup. The system consists of two opposing piezoelectric microgrippers mounted on 3-DOF piezoelectric stages and positioned above a paper surface platform. A silk fiber is grasped at both endpoints and manipulated under top-view visual feedback.

### B. Fiber specimens and manipulation surface

Three silk suture threads of different diameters were used as test fibers: 50 µm, 80 µm, and 120 µm (Fine Science Tools, product codes 18020-70, 18020-60, and 18020-50, Germany). Each fiber was cut slightly longer than required and grasped at both ends such that the manipulated length between the grippers was 10 mm, 15 mm, or 20 mm. These specimens span the range of fiber geometries evaluated in the subsequent experiments. All experiments were performed with the fiber resting on a standard $80\ \mathrm{g/m^2}$ office-paper surface.

### C. Fiber shape control demonstration

We first evaluated the closed-loop behavior in a representative real-world shaping trial using a 15 mm-long silk microfiber with a diameter of 120 µm resting on the paper surface, as shown in Fig. 4(a)-(d). Starting from an initial configuration with a large shape discrepancy, the transferred policy progressively reduces the error between the observed centerline $\mathcal{C}_t$ and the prescribed target $\mathcal{C}^*$ using closed-loop visual feedback. The green line represents the simulated fiber configuration for the same gripper positions, illustrating the mismatch between the friction-free model and the real fiber response.

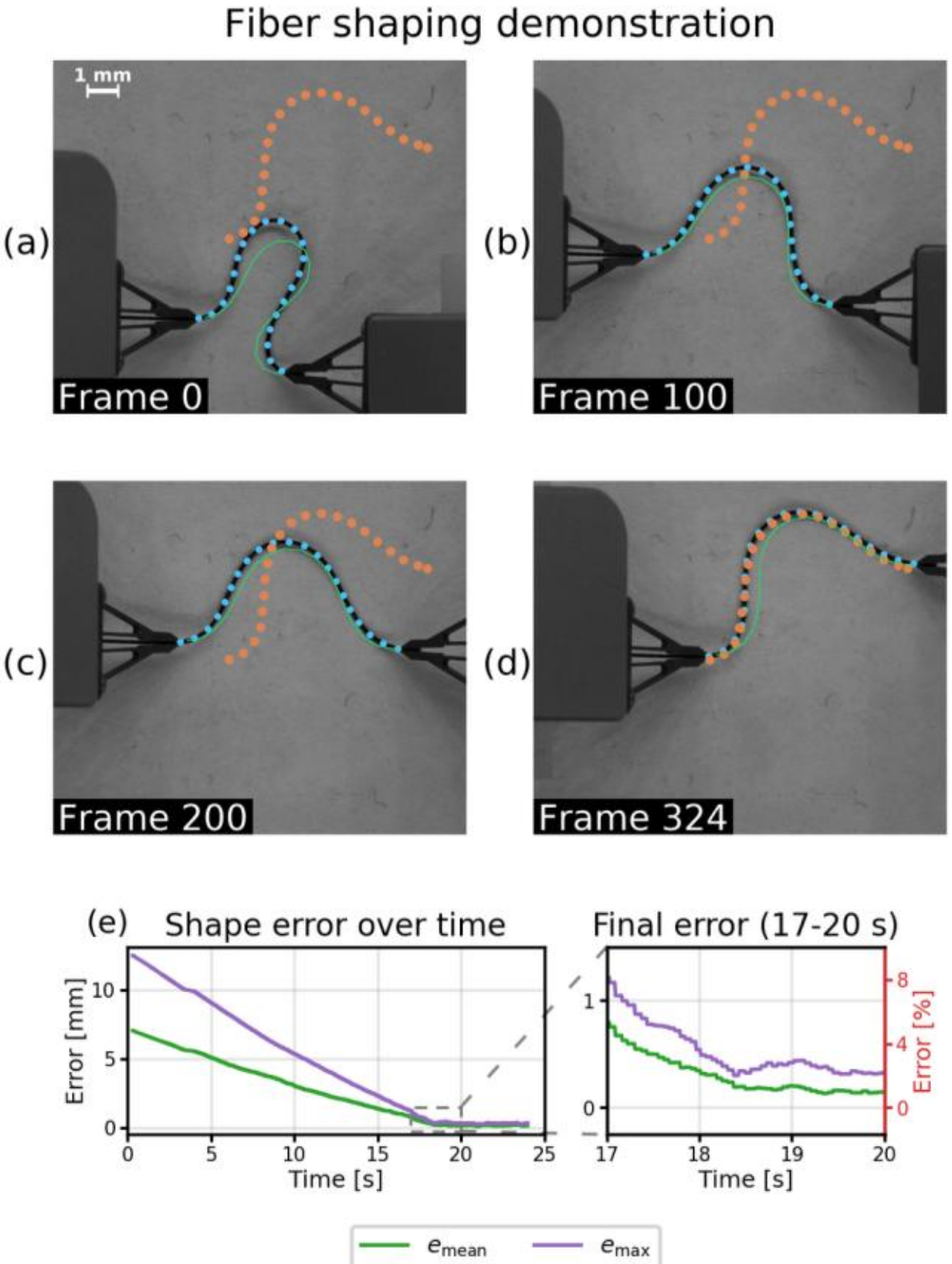

**Fig. 4**: Representative real-world microfiber shaping trial. (a)–(d) Image sequence showing the fiber evolving from the initial configuration to the target shape under closed-loop control. Orange markers indicate the prescribed target points, blue markers indicate the image-derived fiber centerline, and the green line represent the fiber shape in simulation with corresponding grippers position. (e) Evolution of the mean shape error $\boldsymbol{e}_{\mathbf{mean}}$ and maximum shape error $\boldsymbol{e}_{\mathbf{max}}$ during the trial. The red y-axis shows the normalized error, expressed as a percentage of the total fiber length.

In Fig. 4(e), both the mean shape error ($e_{\text{mean}}$) and maximum shape error ($e_{\text{max}}$) decrease during the initial correction phase and then settle into a steady-state regime after ~18 s with final values of $e_{\text{mean}} \approx 160\,\mu\text{m}$, and $e_{\text{max}} \approx 320\,\mu\text{m}$. This example illustrates closed-loop real-world shape regulation under surface contact. The corresponding experimental demonstration, together with additional examples, is provided in Video 1. The following subsections present quantitative results obtained by repeating the same closed-loop shaping procedure across different initial configurations, target shapes, and fiber specimens.

### D. Repeatability across initial configurations

We evaluated repeatability by testing whether the same transferred policy could reach each target shape from different initial fiber configurations. Using a 15mm-long, 50 μm-diameter silk microfiber, we tested three target shapes with distinct curvature profiles, as shown in Fig. 5(a). For each target, eight different initial configurations were generated through teleoperation, yielding 24 trials in total. In each trial, the policy controlled the grippers in a closed loop at 40 Hz until convergence or timeout.

Across the 24 trials, the final fiber shapes converged consistently toward the prescribed targets despite large variations in the starting configurations. The policy achieved a mean point-wise error of $e_{\text{mean}} = 270 \pm 80\,\mu\text{m}$, and a maximum error of $e_{\text{max}} = 390 \pm 100\,\mu\text{m}$. This repeatability across initial states supports the reactive-regulation view: in the tested regime, the controller appears able to rely on currently observed shape deviation without explicitly modeling the contact dynamics or inferring the history dependent trajectory from the initial configuration to the target shape.

The trained policy achieves below 100 μm shape error in simulation, whereas the mean real-world error is 270±80 μm. This difference is consistent with the expected sim-to-real mismatch, including image-based observation noise, unmodeled fiber-surface interactions, and simplifications in the simulated fiber model and physics. Since these effects were not measured independently, their individual contributions cannot be isolated from the present experiments.

### E. Generalization across fiber specimens

To evaluate generalization across physical specimens, we applied the same trained policy, without retraining, to nine silk fibers spanning three diameters (50 μm, 80 μm, and 120 μm) and three manipulated lengths (10 mm, 15 mm, and 20 mm), where 15 mm is the length used during policy training. For each specimen, five target shapes were independently recorded via teleoperation after gripping the fiber, distinct from any shape used during simulation training. Targets were selected to span a range of curvature profiles and degrees of asymmetry, covering configurations representative of the reachable workspace under the tested gripper conditions. Each shape was used in turn as a target while the remaining four shapes served as initial configurations, yielding 20 trials per specimen and 180 trials in total. The targets were selected in a low-complexity regime, comparable to the simpler shapes in Section IV.D, so that differences in shaping accuracy across specimens could be attributed primarily to fiber geometry and physical properties rather than target complexity. The target-complexity distribution is quantified in Section IV.F.

The maximum shape error ($e_{\text{max}}$) across all nine specimen conditions is shown in Fig. 5(b). At every tested length, the 120 μm fibers achieved the lowest median error, while the 80 μm fibers produced the largest, and the 50 μm fibers showed intermediate performance throughout, with overall medians of 251 μm, 414 μm, and 322 μm, respectively. This suggests a systematic effect of fiber diameter on shaping accuracy. Across fiber lengths, the 15 mm training condition yielded the lowest median error of 202 μm, while the 10 mm and 20 mm fibers produced higher medians of 373 μm and 334 μm, respectively. Across all nine conditions, the median error remained below 500 μm, indicating successful policy transfer across the full range of tested specimen geometries without additional training or tuning.

The observed trends with fiber diameter and length may reflect a changing balance between fiber elasticity and surface interaction. Regarding fiber length, the error increased as the fiber length moved away from the 15 mm training condition, suggesting reduced accuracy for fiber geometries farther from the training distribution. With respect to diameter, the 80 μm fibers may lie in an intermediate regime where elastic loading increases contact variability without fully dominating frictional effects, whereas the 120 μm fibers may be less affected by local frictional variations due to their larger restoring forces. This interpretation remains hypothetical because contact forces were not measured directly.

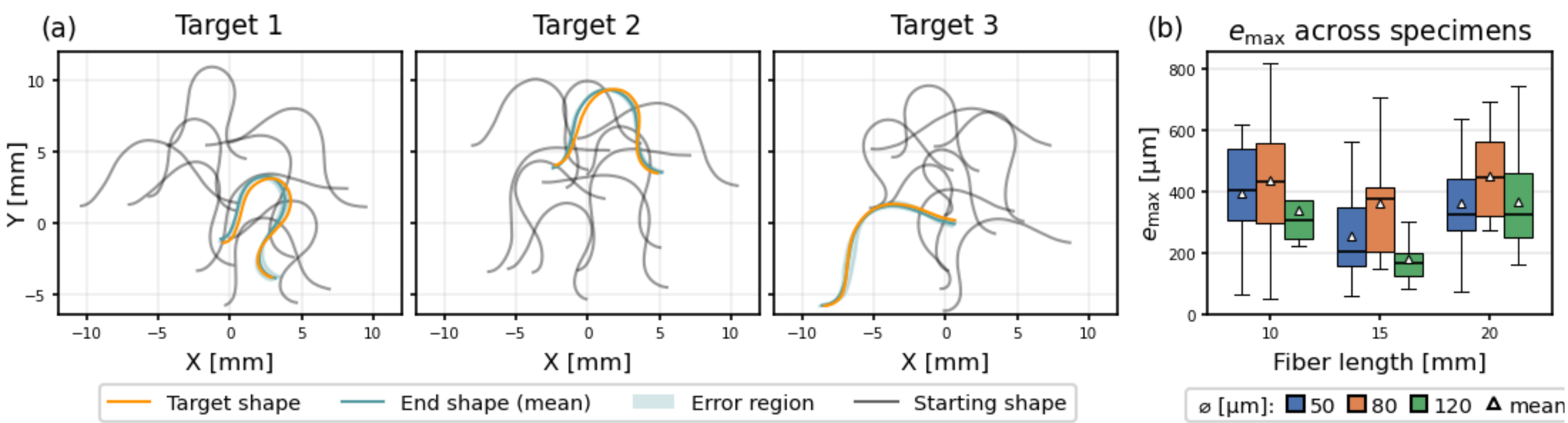

**Fig. 5** Aggregate real-word shaping results. **(a)** Repeatability across initial configurations for three target shapes. Gray curves show initial configurations, orange curves show target shapes, and teal curves show final achieved shapes with error bounds. **(b)** Generalization across fiber specimens of different diameters and lengths. Boxplots of $\boldsymbol{e}_{\textbf{max}}$ per diameter and length; the white triangle marks the mean.

### F. Effect of target geometry

To examine whether target geometry affects shaping accuracy, we characterized each target shape using the bending energy $E_{\mathrm{bend}} = \frac{1}{L}\int_0^L \kappa(s)^2 ds$ [31], defined as the arc-length-averaged integral of squared curvature along the fiber centerline and discretized as $E_{\mathrm{bend}} \approx \frac{1}{L}\sum_i \kappa_i^2 \Delta s_i$. The mean shaping error $e_{\mathrm{mean}}$ was then plotted as a function of $E_{\mathrm{bend}}$ across the evaluated targets, as shown in Fig. 6(a). A positive trend is observed across all diameter groups, suggesting that geometrically more complex targets are generally harder to control accurately.

To compare the geometric complexity of the real experiments with the simulation training set, we also examined the distribution of $E_{\mathrm{bend}}$ for both groups, as shown in Fig. 6(b). The experimental targets occupy a higher bending-energy range than the training distribution, indicating that the policy is evaluated on target geometries that are, under this scalar measure, more complex than those encountered during training. While not definitive on its own, this result suggests that the trained policy can generalize beyond the simpler geometric regime represented in the simulator dataset.

The gradual degradation in accuracy with increasing bending energy, rather than abrupt failure, is consistent with the reactive-regulation view. The feedback loop remained stable even for the most complex targets, producing final fiber configurations that closely matched the prescribed shapes, with mean $e_{\max}$ error below 500 µm across all the tested fibers. The larger residual errors for high-bending-energy targets suggest that final shape correction becomes more difficult as the target geometry becomes more curved or multi-inflectional.

## V. Conclusion

This work presented a closed-loop reinforcement learning approach for microfiber shape control under surface contact. A policy for endpoint-based geometric shape regulation was trained in a simplified frictionless simulator and transferred directly to a physical dual-gripper micromanipulation system, with no retraining or contact-domain randomization. Real-time visual feedback of the fiber centerline was used to iteratively correct shape deviations during manipulation, without any explicit model of fiber-surface friction.

The approach was evaluated on silk microfibers using the same dual-gripper micromanipulation system operating at 40 Hz. A mean point-wise shape error of 270 ± 80 µm was achieved across twenty-four diverse initial configurations. Generalization experiments across nine specimens of varying diameter and length confirmed sub-millimeter final shape error throughout the test range. The experimental targets spanned a higher bending-energy range than the simulated training shapes, indicating that the policy extends to geometrically more complex configurations under this measure.

The approach relies on the assumption that the task-relevant effects of unmodeled contact dynamics are reflected in the observed fiber geometry within the feedback horizon of the controller. This assumption may break down under high-speed manipulation, significant viscoelastic relaxation, or friction conditions substantially different from those encountered during policy evaluation. Future work will investigate such regimes, extending the approach toward faster manipulation, more time-dependent materials, and varying contact surface conditions.

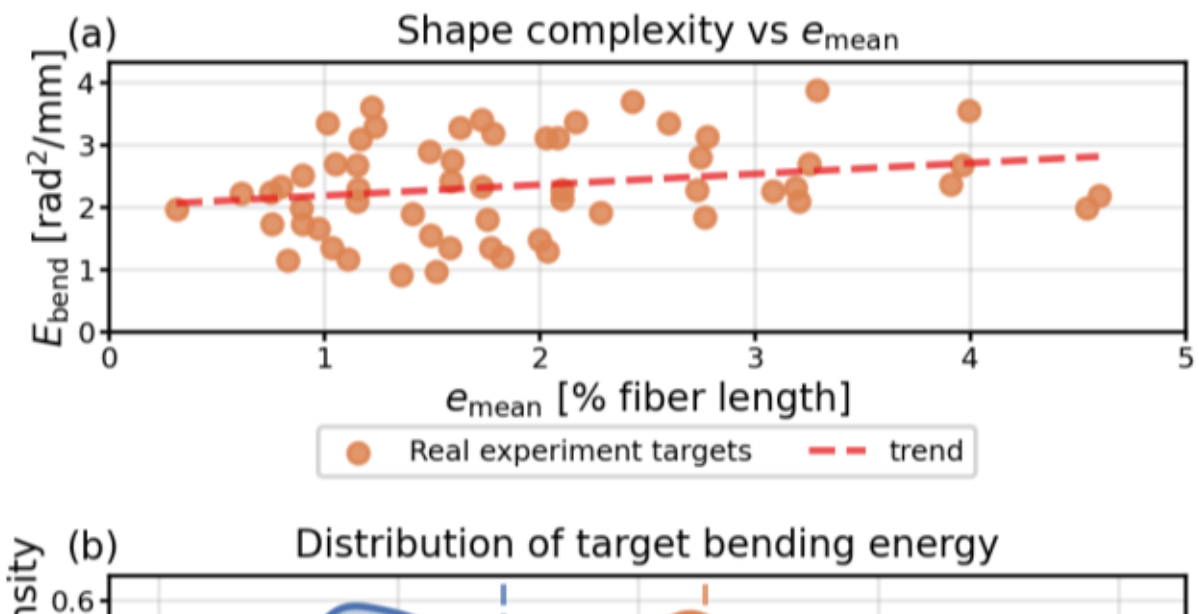

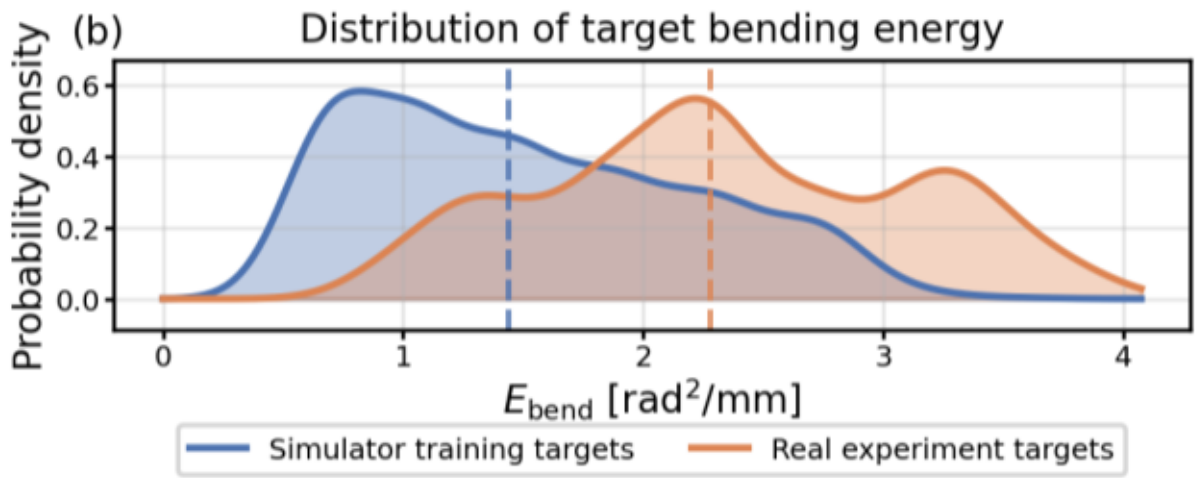

**Fig. 6:** Effect of target bending energy on shape accuracy. **(a)** Mean shaping error $e_{\mathrm{mean}}$ as a function of $E_{\mathrm{bend}}$ across all generalization experiments. **(b)** Probability distribution of $E_{\mathrm{bend}}$ for the simulator training shapes and the real experimental targets; dashed lines indicate the respective medians.

## Appendix

### RL model training performances

Fig. 7 shows the RL training performance over 20 million environment steps across 30 parallel frictionless MuJoCo environments. The mean episode reward increases rapidly during the early training phase and stabilizes after approximately 2 million timesteps, indicating stable training behavior. Consistently, the simulation shape error decreases sharply and reaches values below 0.1 mm, showing that the agent successfully learns the geometric shaping task in the simulator. These results confirm training convergence in the simplified frictionless environment.

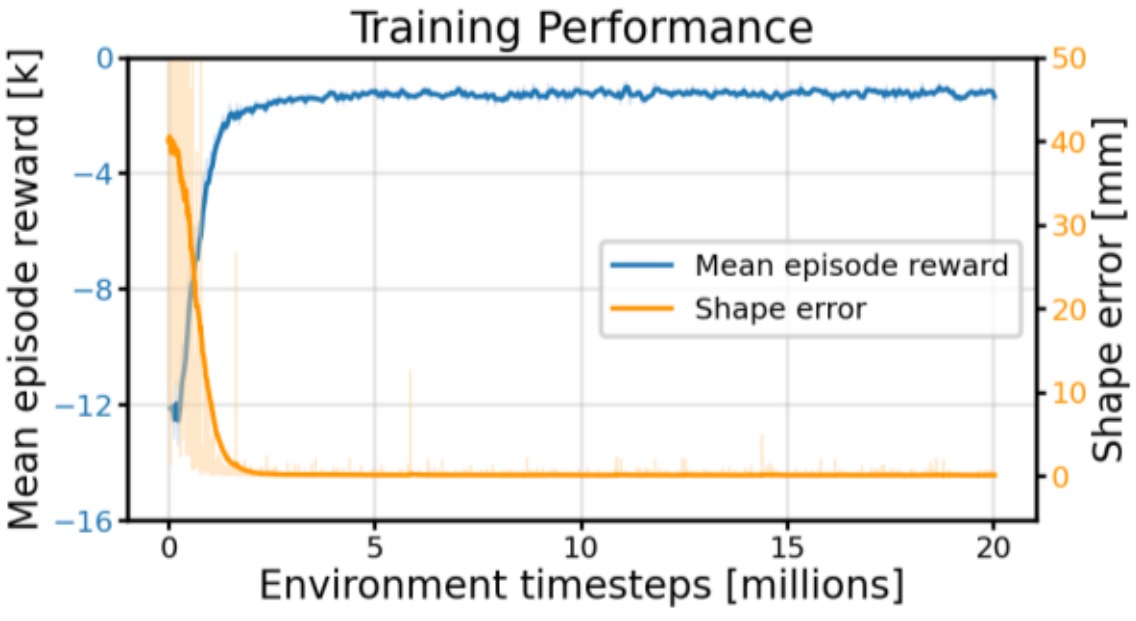

**Fig. 7:** Training performance of the PPO agent over 20 million timesteps. The mean episode reward (blue, left axis) and shape error $\boldsymbol{\epsilon}(\boldsymbol{t})$ (orange, right axis) are shown. The agent converges with low shape error (less than 100 µm) while reward stabilizes, indicating successful training.

## Acknowledgment

We acknowledge the use of Claude.ai (https://claude.ai/) for language improvements. All revisions were thoroughly verified by the authors to preserve the original content.